\documentclass[runningheads,table,xcdraw]{llncs}

\usepackage[T1]{fontenc}
\usepackage{graphicx}

\usepackage[nolist]{acronym}
\usepackage{hyperref}
\usepackage{svg}
\usepackage{array}
\usepackage{amsmath}
\usepackage{amssymb, mathtools}
\usepackage{diagbox}
\usepackage{listings}
\usepackage{subcaption}  %
\usepackage{float}
\usepackage{enumitem}

\usepackage{csquotes}
\MakeOuterQuote{"}

\definecolor{codegreen}{rgb}{0,0.6,0}
\definecolor{codegray}{rgb}{0.5,0.5,0.5}
\definecolor{codepurple}{rgb}{0.58,0,0.82}
\definecolor{backcolour}{rgb}{0.95,0.95,0.92}

\lstdefinestyle{mystyle}{
    backgroundcolor=\color{backcolour},
    commentstyle=\color{codegreen},
    keywordstyle=\color{magenta},
    numberstyle=\tiny\color{codegray},
    stringstyle=\color{codepurple},
    basicstyle=\ttfamily\footnotesize,
    breakatwhitespace=false,
    breaklines=true,
    captionpos=b,
    keepspaces=true,
    numbers=left,
    numbersep=5pt,
    showspaces=false,
    showstringspaces=false,
    showtabs=false,
    tabsize=2
}

\begin{document}
\newcommand{\authorA}{Klara M. Gutekunst}
\newcommand{\authorB}{Dominik Dürrschnabel}
\newcommand{\authorC}{Johannes Hirth}
\newcommand{\authorD}{Gerd Stumme}

\newcommand{\emailA}{klara.gutekunst@uni-kassel.de}
\title{Conceptual Topic Aggregation}
\titlerunning{Conceptual Topic Aggregation}
\author{\authorA{}\orcidID{0009-0002-6416-7873}
        \and \authorB{}\orcidID{0000-0002-0855-4185}
        \and \authorC{}\orcidID{0000-0001-9034-0321}
        \and \authorD{}\orcidID{0000-0002-0570-7908}
}

\authorrunning{\authorA{}
        \and \authorB{}
        \and \authorC{}
        \and \authorD{}
}

\institute{Knowledge \& Data Engineering Group, University of Kassel, Germany
        \and Interdisciplinary Research Center for Information System Design,
        \mbox{University of Kassel, Germany} \\%
        \email{\emailA{}\\ \{duerrschnabel, hirth, stumme\}@cs.uni-kassel.de}}
\maketitle              %

\begin{abstract}    %

    The vast growth of data has rendered traditional manual inspection infeasible, necessitating the adoption of computational methods for efficient data exploration. Topic modeling has emerged as a powerful tool for analyzing large-scale textual datasets, enabling the extraction of latent semantic structures. However, existing methods for topic modeling often struggle to provide interpretable representations that facilitate deeper insights into data structure and content.

    In this paper, we propose \acs{approachname}, an approach based on \ac{fca} to enhance meaningful topic aggregation and visualization of discovered topics. Our approach can handle diverse topics and file types -- grouped by directories -- to construct a concept lattice that offers a structured, hierarchical representation of their topic distribution.

    In a case study on the ETYNTKE dataset, we evaluate the effectiveness of our approach against other representation methods to demonstrate that \acs{fca}-based aggregation provides more meaningful and interpretable insights into dataset composition than existing topic modeling techniques.

\end{abstract}

\section{Introduction}
\label{sec:introduction}

Investigating large, unstructured datasets with diverse file types is a common challenge, especially in domains such as tax investigation. When leaks potentially involving tax fraud emerge, human analysts must sift through unknown data to find relevant evidence. To support this task, we propose using concept lattices to structure the data hierarchically, enabling efficient identification of meaningful topics and patterns.

Current methods such as LDA, BERTopic, and Top2Vec can be used for the identification of topics in documents.
However, these techniques do not leverage the intrinsic hierarchical structure of the topics encapsulated in the documents.
One attempt in the realm of \ac{fca} \cite{topic_modeling_2024}  is to overcome this shortcoming by  creating document-topic relations and use the concept lattice as a hierarchical representation.
However, as they work on independent and homogeneous documents, which are exclusively of textual nature,
this method lacks the ability to aggregate information.

In this work, we introduce \ac{approachname},
which integrates \ac{nlp} with \ac{fca},
leveraging their complementary strengths to structure and visualize relationships within data.
Thereby, we can incorporate heterogenous data, consisting of different data types and content.
The main contributions of this work are:
\begin{enumerate}[label=(\alph*)]
    \item Our approach integrates non-textual image data.
    \item We employ \ac{fca} to aggregate topic information effectively, facilitating insight combination across multiple directories.
    \item \ac{approachname} is designed to operate with minimal training requirements, utilizing pre-trained, off-the-shelf models and thereby removing the necessity for extensive training.
    \item We demonstrate through a case study on a real-world dataset, how the approach can support a human analyst.
    \item We provide an open-source implementation, accessible on GitHub\footnote{\url{https://github.com/KlaraGtknst/text_topic} (29.01.2025) \&\\ \url{https://github.com/KlaraGtknst/clj_exploration_leaks} (29.01.2025)}.
\end{enumerate}

The structure of this paper is as follows:
In Section~\ref{sec:related_work}, we review relevant literature.
Section~\ref{sec:foundations} presents foundations, followed by Section~\ref{sec:method},
which details our proposed methodology \ac{approachname}.
In Section~\ref{sec:baselines}, we present three baseline strategies.
Section~\ref{sec:implementation} outlines implementation details of our approach.
In Section~\ref{sec:case_study}, we present a case study on a real-world dataset, including results and an empirical assessment of our approach alongside the baselines.
Finally, we conclude with Section~\ref{sec:future_work} and outline potential directions for future research.

\section{Related Work}
\label{sec:related_work}

The authors of \cite{twitter_fca_2016} apply \ac{fca} to a Twitter dataset to extract insights into the topics present.
In their methodology, tweets serve as objects, while words occurring within the tweets act as attributes.
They assume that tweets forming a formal concept, i.e., those sharing the same terminology, correspond to the same topic.
Upper concepts in the resulting concept lattice represent general topics,
whereas lower-level concepts reflect more specific themes.
However, a key limitation of their approach is the excessive number of concepts generated, making interpretation challenging.
To mitigate this, they introduce upper and lower frequency thresholds,
discarding terms whose occurrence falls outside these predefined bounds,
arguing that the former are too general and the latter too specific for meaningful analysis.
While their approach provides valuable means as how to derive topic structures from text data,
it relies purely on term frequency as attributes rather than
leveraging the semantic relationships inherent in topic models, as we do in this work.

The authors of \cite{topic_feature_2021} employ \ac{fca} to structure topic representations derived from a topic model.
They construct two matrices:
One capturing document-topic probabilities obtained through their ILDA model and
another representing topic-word probabilities, restricted to the top ten words per topic.
Unlike our approach, which automatically derives hierarchical relationships,
their method relies on manual categorization of topics into broader thematic groups.
These manually curated categories are subsequently visualized using a \ac{fca}-inspired order diagrams to reflect topic generalization.
In contrast, our approach seeks to infer hierarchical topic structures automatically, without requiring manual intervention.

According to \cite{HLDA_2019}, \ac{hlda} organizes extracted topics into a tree,
and requires documents with clear structure and rigorous style.
Since our approach aims to analyze the structure of diverse data rather than demanding a specific structure,
we do not employ \ac{hlda}.
Moreover, we chose to employ \ac{fca} since trees give less structural information than lattices.

The authors of \cite{topic_hierarchy_2023} also leverage \ac{fca} to structure topic hierarchies.
They utilize LDA for topic modeling, encoding documents based on the distribution of topics.
Similar to our approach, they introduce a thresholding mechanism to retain only relevant topics per document.
However, a major divergence lies in their reliance on HowNet, a linguistic knowledge base.
Their primary objective is to facilitate precise querying,
using a concept lattice to provide a more nuanced representation of documents beyond mere word-level indexing.
To manage the complexity of the lattice, they eliminate concepts whose attribute sets exhibit excessively high or low support.
While effective for information retrieval, their approach differs from ours in fundamental ways.
The focus of our approach is uncovering hierarchical relationships within topic structures across the dataset,
rather than enhancing document retrieval.
We do not discard high-support concepts,
as they provide crucial insights into overarching themes.

Our approach builds on the work of the authors of \cite{topic_modeling_2024} who propose extracting two relational structures from topic models.
The first structure is a weighted document-topic relation, i.e., which topics $t_i$ are present in which documents $d_j$.
The second structure is a weighted topic-word relation, i.e., which words $w_k$ are present in which topics $t_i$.
These relations are subsequently used to derive incidence matrices,
as visualized in Tables \ref{tab:doc-topic} and \ref{tab:term-topic}.

\begin{table}[t]
    \centering
    \begin{minipage}{0.5275\textwidth}
        \centering
        \caption{Document-topic relation}
        \label{tab:doc-topic}
        \begin{tabular}{|c|c|c|c|c|}
            \hline
            \backslashbox{Documents}{Topics} & $t_1$     & $t_2$     & $\cdots$ & $t_n$     \\ \hline
            $d_1$                            & $w_{1,1}$ & $w_{1,2}$ & $\cdots$ & $w_{1,n}$ \\ \hline
            $d_2$                            & $w_{2,1}$ & $w_{2,2}$ & $\cdots$ & $w_{2,n}$ \\ \hline
            $\vdots$                         & $\vdots$  & $\vdots$  & $\ddots$ & $\vdots$  \\ \hline
            $d_l$                            & $w_{l,1}$ & $w_{l,2}$ & $\cdots$ & $w_{l,n}$ \\ \hline
        \end{tabular}
    \end{minipage}%
    \hfill
    \begin{minipage}{0.465\textwidth}
        \centering
        \caption{Term-topic relation}
        \label{tab:term-topic}
        \begin{tabular}{|c|c|c|c|c|}
            \hline
            \backslashbox{Terms}{Topics} & $t_1$           & $t_2$           & $\cdots$ & $t_n$           \\ \hline
            $s_1$                        & $\hat{w}_{1,1}$ & $\hat{w}_{1,2}$ & $\cdots$ & $\hat{w}_{1,n}$ \\ \hline
            $s_2$                        & $\hat{w}_{2,1}$ & $\hat{w}_{2,2}$ & $\cdots$ & $\hat{w}_{2,n}$ \\ \hline
            $\vdots$                     & $\vdots$        & $\vdots$        & $\ddots$ & $\vdots$        \\ \hline
            $s_l$                        & $\hat{w}_{l,1}$ & $\hat{w}_{l,2}$ & $\cdots$ & $\hat{w}_{l,n}$ \\ \hline
        \end{tabular}
    \end{minipage}
\end{table}

To compute a binary relation from the document-topic relation, they apply a thresholding approach,
while for the term-topic relation, they retain only the top $n$ words per topic.
Subsequently, the authors compute the formal concepts of high support.
Their approach enables a structured representation of topic hierarchies,
which they demonstrate on academic papers from machine learning conferences.
The approach lacks the possibility to compare subsets of the whole dataset, which might stem from an organization of files into directories.
By contrast, our approach addresses this limitation by explicitly incorporating cross-directory topic structures,
allowing for a more comprehensive analysis of thematic relationships spanning multiple data sources.
Furthermore, our method is not restricted to only textual data.

\section{Foundations}
\label{sec:foundations}

\paragraph{\acf{fca}} The mathematical theory of \ac{fca}~\cite{fca-book} is a method to analyze relational data in a hierarchical structure.
The triple $\mathbb{K} \coloneq (G, M, I)$ defines the \emph{formal context}.
$G$ is the set of \emph{objects}, $M$ is the set of \emph{attributes},
and $I \subseteq G \times M$ is the \emph{incidence relation}, which describes when an object $g$ has an attribute $m$.
The object derivation $(\cdot)' : \mathcal{P}(G) \to \mathcal{P}(M)$
defines for any subset $A \subseteq G $ as $A' \coloneq \{m \in M \mid \forall g \in A : (g, m) \in I\}$.
The attribute derivation $(\cdot)' : \mathcal{P}(M) \to \mathcal{P}(G)$ is
defined for any subset $B \subseteq M $ as $B' \coloneq \{g \in G \mid \forall m \in B : (g, m) \in I\}$.
The same notation $(\cdot)'$ is used for both derivation operators.
A formal concept of $\mathbb{K}$ is a pair $(A, B)$ where $A \subseteq G$ and $B \subseteq M$
satisfy the conditions $A' = B$ and $B' = A$.
The set $\mathfrak{B}(\mathbb{K})$ denotes all formal concepts of $\mathbb{K}$.
The concept lattice $\underline{\mathfrak{B}}(\mathbb{K}) \coloneq (\mathfrak{B}(\mathbb{K}), \leq)$ is defined by the set of all formal concepts with the order relation
$(A, B) \leq (C, D)$ if and only if $A \subseteq C$.

\paragraph{Image Captioner}

An image captioner is a tool to generate textual descriptions from images.
In this paper, we specifically employ the
\ac{git}\footnote{\url{https://huggingface.co/microsoft/git-base} (29.01.2025)}
model \cite{git_2022}.
It is a straightforward generative model comprising an image encoder and a text decoder.
The image encoder is a Contrastive Learning-pretrained, Swim-like vision transformer,
which outputs a flattened 2D representation of the image.
This vector is then projected into D dimensions before being passed to the text decoder,
a transformer module, which generates coherent textual descriptions.
This pre-trained model has the ability to generate meaningful captions without the need for custom training,
enabling us to seamlessly integrate visual information into our topic modeling process.

\paragraph{Top2Vec}

The pre-trained topic model Top2Vec~\cite{top2vec_2020}, which is  employed in this study, embeds texts into the same vector space as words and topics.
This is particularly advantageous,
as it allows for semantic similarity calculations between texts, words, and topics.
The model encodes texts using a multilingual \ac{use} model \cite{use_2019}, supporting 16 languages,
which is crucial for working with diverse datasets containing multiple languages.
The documents are first embedded into a 300-dimensional space, then reduced to 5 dimensions and finally clustered into topics.
Due to the modular design of our approach,
it is straightforward to substitute Top2Vec with alternative topic modeling algorithms,
such as BERTopic, that offer comparable functionality.

\paragraph{TITANIC}

Computing concept lattices for large datasets presents a significant challenge, as the size of a concept lattice can grow exponentially with respect
to the size of the context.
Consequently, complete visualizations of large concept lattices may obscure important hierarchical
relationships within the data, hindering human interpretation.
To address this issue, \cite{titanic_2002} introduced the TITANIC algorithm,
which computes iceberg concept lattices.
An iceberg concept lattice includes only the top-level concepts,
specifically those corresponding to frequent attribute sets.

The authors define a minimum support threshold,
$minsupp \in [0,1]$, below which attribute sets are not considered frequent.
An attribute set $B \subseteq M$ is deemed frequent in $\mathbb{K}$ if its support
$$supp(B) = \frac{\left| B' \right|}{\left| G \right|}$$ meets or exceeds the $minsupp$ threshold.
A concept $(A, B)$ is considered frequent if its intent $B$ constitutes a frequent attribute set.
The iceberg concept lattice of a context $\mathbb{K}$ comprises all such frequent concepts.

\section{Method}
\label{sec:method}

\begin{figure}[t]
    \includesvg[width=\textwidth]{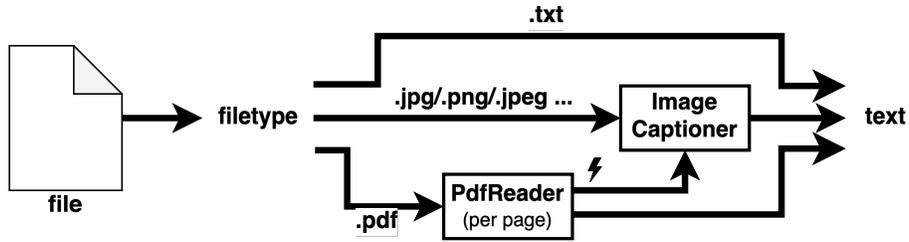}
    \caption{Visual representation of the workflow for extracting text from the data files.
        If a page of a PDF document contains no text or raises an error during extraction,
        the image captioner is used to generate a textual description of the image of that page.}
    \label{fig:extract_text}
\end{figure}

The \emph{\acf{approachname}} methodology introduced in this paper  outlines a systematic workflow for
deriving the directory-topic concept lattice of a dataset.
First, we extract the text from the data files.
The processing of text-related fields follows a distinct workflow, as shown in Figure \ref{fig:extract_text}.

\begin{figure}[b]
    \centering
    \includegraphics[width=0.5\textwidth]{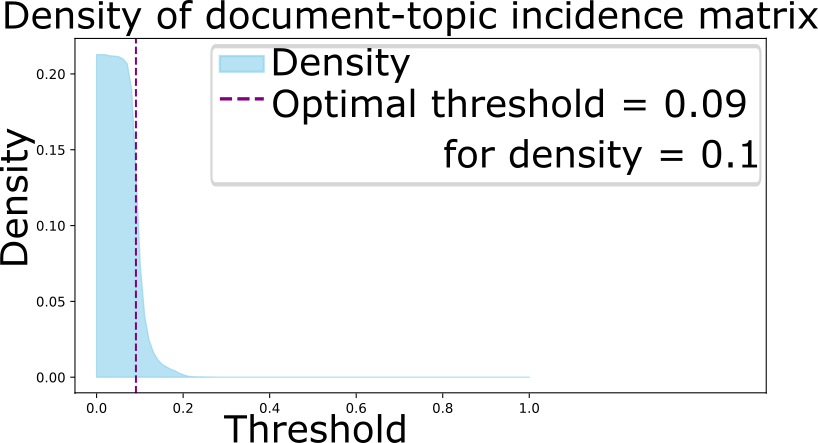}
    \caption{Computation of threshold $\delta$ such that the resulting thresholded binary document-topic incidence $\mathcal{D}_\delta$ has a density of $\frac{\left| \mathcal{D}_\delta \right|}{\left| D \times T \right|}=0.1$.}
    \label{fig:fca_threshold}
\end{figure}

By employing a topic model, specifically the Top2Vec model,
we extract the topics of the documents.
While the model typically returns only one topic per document,
we retrieve the top ten topics for each document to capture a broader range of thematic content.
Additionally, the model provides topic scores, topic words, and word scores,
which we leverage to derive document-topic relationships, as proposed by \cite{topic_modeling_2024} (cf.~Table \ref{tab:doc-topic}).
These topics are assigned real values $w_{i,j} \in \mathbb{R}$,
which indicate the relevance of a respective topic $t_i$ to document $d_j$.
Next, these weights are row-normalized.
The binary document-topic incidence matrix is then created by applying a threshold $\delta$,
which helps in determining the most relevant topics to each document.
If $w_{i,j} \ge \delta$, topic $t_i$ is present in document $d_j$.
The threshold is the first value such that the density of thresholded weights is below a certain density threshold.
With accordance to \cite{topic_modeling_2024}, we set the density threshold to $0.1$
to obtain a reasonably sparse binary incidence
leading to a comprehensibly sized concept lattice including the most important topics.
The computation resulting in $\delta=0.09$ is visualized in Figure \ref{fig:fca_threshold}.
After $\delta$ is determined, we obtain the binary document-topic incidence matrix
$$\mathcal{D}_\delta(i,j)=\left\{ \begin{array}{cl}
        1 & \ \text{if } \ w_{i,j} \geq \delta, \\
        0 & \ \text{else.}
    \end{array} \right.$$
This matrix serves as the foundation for understanding how topics are distributed across different directories.

\begin{figure}[H]
    \centering
    \includegraphics[width=0.9\textwidth]{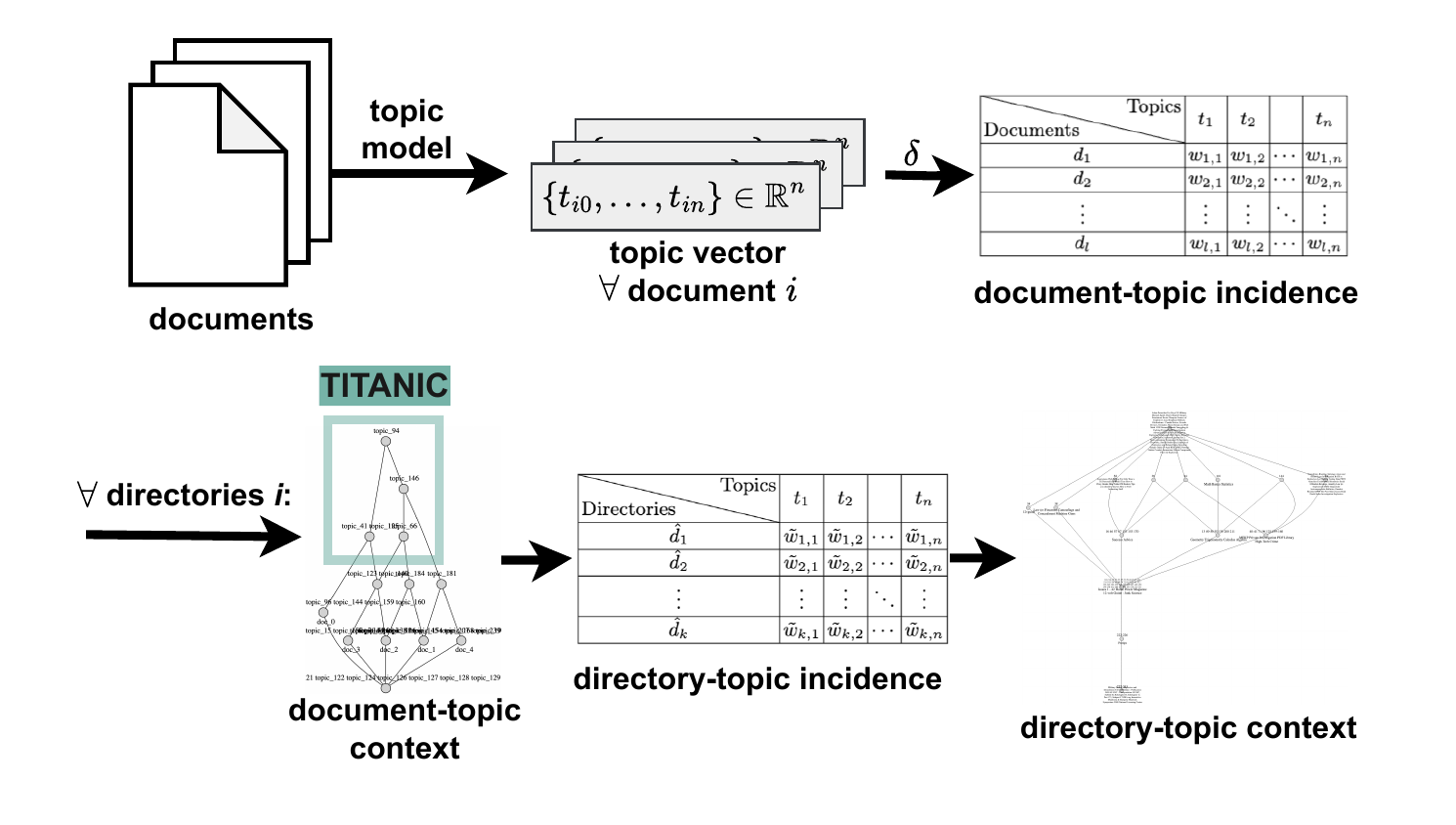}
    \caption{Workflow for deriving the directory-topic concept lattice.}
    \label{fig:fca_dirs_fca}
\end{figure}

Compared to the approach of Hirth and Hanika~\cite{topic_modeling_2024},
we aggregate the binary document-topic incidence matrices from
all directories to gain a broader understanding of the dataset's topic structure as a whole.
The workflow is visualized in Figure \ref{fig:fca_dirs_fca}.

We adopt the TITANIC algorithm
as it generates interpretable concept lattices from large datasets, making it well-suited for our study.
For each directory, we compute an iceberg concept lattice.
We focus on high-support concepts because their attribute sets represent topics
that are consistently present across many documents in a directory.
These frequent topics capture broader themes that are widely distributed throughout the respective directory,
making them ideal for constructing a representative set of attributes.
The minimum support is a parameter that should be chosen according to the degree of detail in which the data ought to be explored.
According to the experience of the authors, a starting parameter of 0.1 is recommendable.
The sequence of frequent concepts over multiple directories is then used to build the incidence matrix
presented in Table \ref{tab:dir-topic}.
This aggregation allows for a hierarchical investigation of the dataset,
where more general topics appear at the top of the hierarchy
and more specific topics are found towards the bottom.

\begin{table}[t]
    \centering
    \caption{Weighted directory-topic relation.
        The weights $\tilde{w}_{i,j} \in \{0,1\}$ represent whether a topic $t_j$ is present in a directory $\hat{d}_i$.}
    \label{tab:dir-topic}
    \begin{tabular}{|c|c|c|c|c|}
        \hline
        \backslashbox{Directories}{Topics} & $t_1$             & $t_2$             &          & $t_n$             \\ \hline
        $\hat{d}_1$                        & $\tilde{w}_{1,1}$ & $\tilde{w}_{1,2}$ & $\cdots$ & $\tilde{w}_{1,n}$ \\ \hline
        $\hat{d}_2$                        & $\tilde{w}_{2,1}$ & $\tilde{w}_{2,2}$ & $\cdots$ & $\tilde{w}_{2,n}$ \\ \hline
        $\vdots$                           & $\vdots$          & $\vdots$          & $\ddots$ & $\vdots$          \\ \hline
        $\hat{d}_k$                        & $\tilde{w}_{k,1}$ & $\tilde{w}_{k,2}$ & $\cdots$ & $\tilde{w}_{k,n}$ \\ \hline
    \end{tabular}
\end{table}

\section{Baselines}
\label{sec:baselines}

The following sections outline common topic visualization strategies serving as baselines for our approach.
We begin with word clouds, a simple, frequency-based visualization baseline of the most common words in a document.
Next, we describe the \ac{edr} approach that clusters low-dimensional embeddings of the documents by directories.
We then explore \ac{cne}.

\paragraph{Word clouds}

An initial approach for visualizing topics involves using word clouds,
a simple and intuitive method for displaying the most frequent words in a document.
A word cloud is a visual representation of text data where words are displayed in varying sizes based on their frequency or importance. It helps identify key themes and trends in textual content at a glance.
Since we are interested in the content of the directories,
we generate word clouds for each directory in the dataset.
The Figures \ref{fig:wordcloud_military} and \ref{fig:wordcloud_spytech}
show examples of word clouds.

\paragraph{\acf{edr}}
\label{sec:low_dim}

\begin{figure}[h]
    \centering
    \includesvg[width=0.7\textwidth]{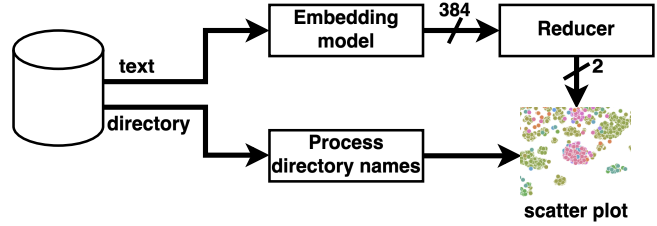}
    \caption{Illustration of the dimensionality reduction process for document embeddings colored by directory labels.
        The embedding model \ac{sbert} omits $384$ dimensional embeddings for each text.
        They are reduced to two dimensions using \ac{pca}, \ac{umap} or \ac{tsne}.
        The directory names are preprocessed to group similar ones and reduce the number of classes since
        they are used to color the points in the scatter plot.}
    \label{fig:low_dim_dir_colour}
\end{figure}

The second baseline approach \ac{edr}, whose workflow is illustrated in Figure \ref{fig:low_dim_dir_colour},
aims to cluster documents by directories
to investigate whether the files within a directory exhibit semantic similarity in terms of content.
To achieve this, the text from the files is embedded with the \ac{sbert} model in a vector space such that distance correlates with semantic dissimilarity.
Subsequently, dimensionality reduction is employed to enable visualization via a scatter plot.
For this baseline we tried the three different dimensionality reducing algorithms
\ac{pca}, \ac{umap}, and \ac{tsne} (cf. Figure \ref{fig:low_dim_dir_colour_tsne}) and settled on \ac{tsne}, as the other two empirically lacked the ability to clearly separate of the data points.
The directory names serve as class labels for the files, i.e., they give the color of the presented points.
To make the method applicable to large-scale datasets,
the directory names are preprocessed to reduce the number of classes.
Specifically, directory names that contain more than $50\%$ numerical characters are replaced with the label "numbers",
while any remaining directory names containing numbers have the numbers stripped.
Directory names with fewer than five alphabetic characters are replaced with the label "chars".

\paragraph{\acf{cne}}

\begin{figure}[!h]
    \includesvg[width=\textwidth]{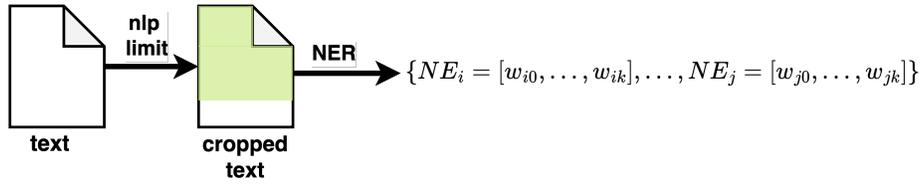}
    \caption{Visual representation of the workflow from text processing to the structure storing \acp{ne} for a given document.}
    \label{fig:ner}
\end{figure}

The baseline \ac{cne} builds on \ac{ner}, which identifies and classifies entities into predefined categories,
such as organizations, locations, dates, and other relevant terms \cite{ner_2019}.
Thereby,
\ac{ner} provides a method for categorizing text. %
The \ac{ner} workflow, as shown in Figure \ref{fig:ner},
begins with truncating the text to meet the input length constraints of the \texttt{nlp} object.
The \ac{ner} model is then applied to the text,
restructuring the identified \acp{ne} so that each entity category functions as a
key mapping to a list of corresponding \acp{ne}.
This structure allows for the efficient retrieval of named entities by category,
facilitating their use in subsequent analyses.

\begin{figure}[H]
    \centering
    \includesvg[width=\textwidth]{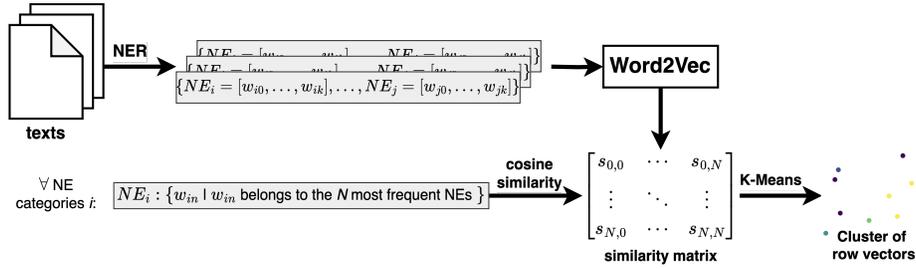}
    \caption{Illustrates the workflow for deriving a clustering of the \ac{ne} categories $i$.}
    \label{fig:ne_clustering}
\end{figure}

In this baseline, we disregard the hierarchical directory structure and
focus on the clustering of \acp{ne} for distinct categories.
The approach of clustering of \acp{ne} has been proposed by \cite{ner_2019}.
The workflow of this baseline is illustrated in Figure \ref{fig:ne_clustering}.
First, \ac{ner} is applied to all texts.
While \cite{ner_2019} replaces similar \acp{ne} with a single version,
referred to as soft entity linking, we omit this step,
though it may improve clustering results.

\begin{figure}[H]
    \centering
    \includesvg[width=0.4\textwidth]{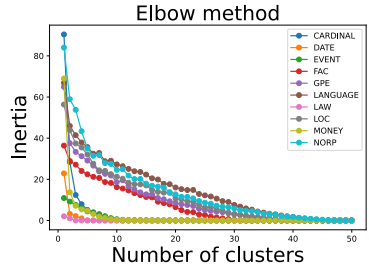}
    \caption{Elbow method for selecting $k$ in $k$-means across selected \ac{ne} categories.}
    \label{fig:elbow_kmeans_cluster_NEs}
\end{figure}

Subsequently, \acp{ne} are embedded using a language model
that produces embeddings comparable via cosine similarity.
This study employs the pre-trained Word2Vec model.
For each \ac{ne} category $i$,
the $N=50$ most frequent \acp{ne} are extracted.
A symmetric $N \times N$ similarity matrix is computed for the most frequent \acp{ne} within each category,
using cosine similarity as a similarity measure.
Each row of the similarity matrix is treated as the feature vector of a given \ac{ne}.
These \acp{ne} are then clustered using the $k$-Means algorithm, with $k=5$.
We chose $k=5$ as a reasonable number of clusters for the \ac{ne} categories,
after consulting the elbow method
across multiple \ac{ne} categories as displayed in \autoref{fig:elbow_kmeans_cluster_NEs}.
Choosing more than five clusters does not substantially increase the compactness of the clusters.
Both $N$ and $k$ are hyperparameters which can be optimized.

\section{Implementation}
\label{sec:implementation}

This section details the implementation of the key components of our approach.
Several \ac{fca} tools used in this work are implemented by the \texttt{conexp-clj}
library~\cite{hanika2019conexp}.
We derive iceberg concept lattices via the TITANIC algorithm.
We choose a minimum support threshold of $0.1$ as we usually observed around 15 concepts which is a reasonable amount to be readable by a human analyst.
For text visualization via word clouds, we utilize the Python library
\texttt{wordcloud}\footnote{\url{https://github.com/amueller/word_cloud} (01.02.2025)}
which is employed with its default parameters.
To extract \acp{ne}, we employ the pre-trained \texttt{en\_core\_web\_sm} pipeline from the spaCy library,
serving as the \texttt{nlp} object.
This lightweight model, trained on English text, provides a general-purpose pipeline for tagging, parsing,
lemmatization, and \ac{ner}\footnote{\url{https://spacy.io/models} (13.02.2025)}.
However, this implementation is constrained to a maximum of $10^6$ tokens\footnote{\url{https://spacy.io/api/language} (30.01.2025)}.

\section{Case Study}
\label{sec:case_study}

To assess the applicability of our approach \ac{approachname},
we apply our method alongside three baselines to the \acs{dataset} dataset.
The next section describes the dataset,
followed by an evaluation of our approach against baseline techniques.

\subsection{Dataset}
\label{sec:data}

The dataset utilized in this study, referred to as \ac{dataset}\footnote{\url{https://archive.org/details/ETYNTKE} (29.01.2025)},
comprises approximately 101 GB of images, text documents, and other file types uploaded in 2015.
It covers a diverse range of topics, including politics, weaponry, and computer science,
providing a rich and varied foundation for extracting semantic insights.
This diversity is particularly suited for the application of our \ac{fca} approach,
as it enables the exploration of patterns across multiple domains, showcasing the versatility of the methodology.

Structurally, the dataset follows a hierarchical organization, with files arranged in folders and subfolders.
The presence of malformed files, which cannot be opened or read,
adds an additional layer of complexity,
making the dataset an ideal challenge for testing the robustness and adaptability of our approach.

The large size, heterogeneity, and hierarchical structure of \ac{dataset}
make it an ideal candidate for applying our \ac{approachname} methodology,
as we can leverage the organizational structure in our approach.
Its variety of content, along with the potential inconsistencies in its organization,
provides a comprehensive testbed for exploring how \ac{nlp} and \ac{fca}
can work together to extract meaningful insights and structure complex datasets.

\subsection{\ac{approachname}}

\begin{figure}[t]
    \centering
    \includegraphics[width=\textwidth]{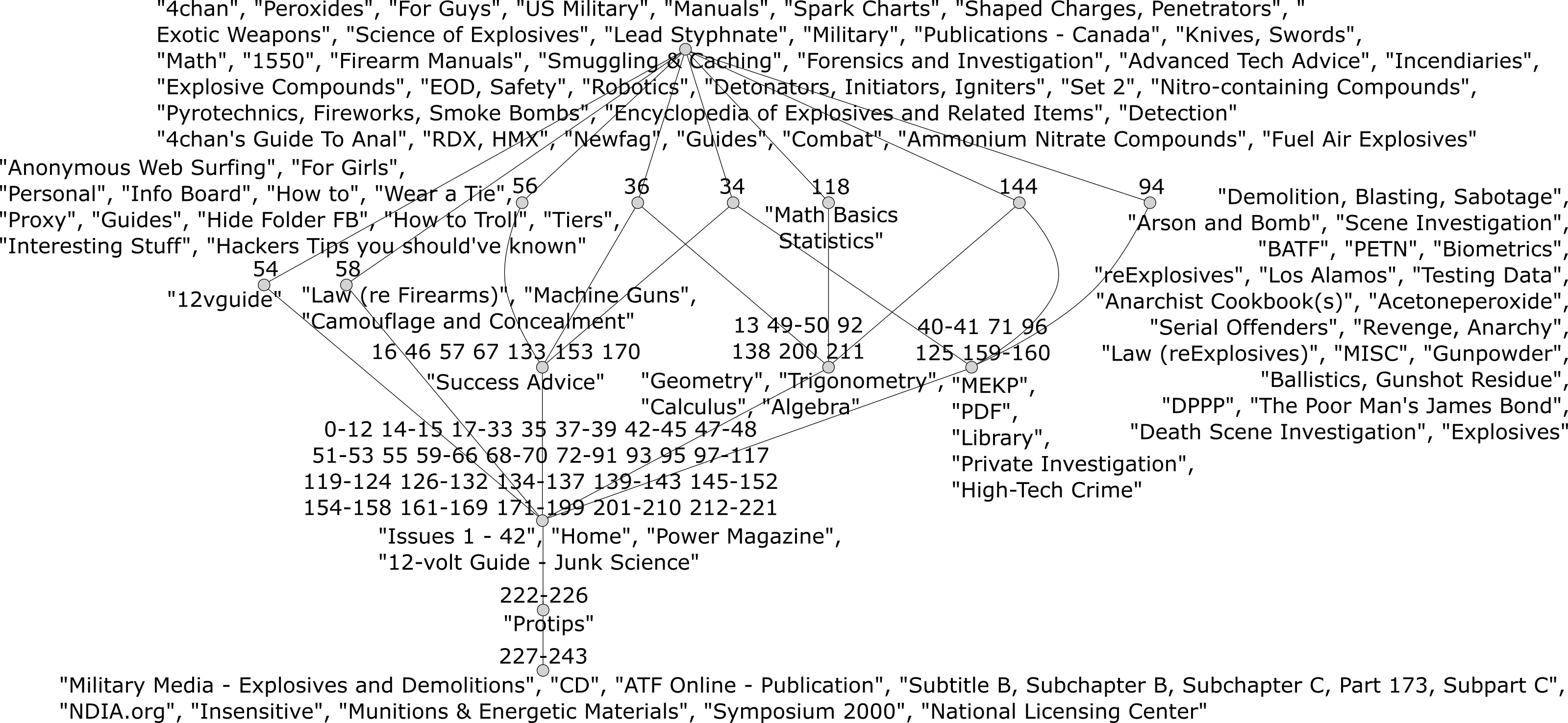}
    \caption{Concept lattice representing the directory-topic incidence of the dataset \ac{dataset}.
        The numbers correspond to topic IDs, with each topic characterized by 50 words, as determined by the Top2Vec model.
        A subset of the topic IDs is mapped to corresponding terms in Table \ref{tab:dir_topic_topics2terms}.
        The text entries below the nodes represent directory names.
    }
    \label{fig:fca_across_dirs}
\end{figure}

The lattice resulting from the \ac{dataset} dataset is illustrated in Figure \ref{fig:fca_across_dirs}.
Topics, represented by numbers above the nodes,
propagate downward to transitive adjacent nodes.
Consequently, every node connected to and below node 94 also contains topic 94.
Topics associated with higher nodes are more common among directories,
whereas lower nodes contain more specific topics found in fewer directories.
The uppermost node represents topics shared across all directories.
In this case, no such topic exists.
Ranges of consecutive topic numbers are denoted by their lowest and highest IDs,
separated by a hyphen.
A subset of topic words for a selection of topic IDs is given in Table \ref{tab:dir_topic_topics2terms}.
The topic model successfully merged content from different languages into single, unified topics, indicating it captures cross-lingual semantic similarity. The concept lattice then organizes these topics by structural relationships, enabling a hierarchical investigation of the dataset through the ordering of concepts. While the topic model extracts and clusters semantic content, the concept lattice structures and contextualizes these clusters to support exploratory analysis.
As directories propagate upwards, lower nodes correspond to fewer directories.

A motif is a statistically significant subgraph or pattern \cite{hirth_ordinal_2024}.
The presence of such structural patterns, which can be identified in Figure \ref{fig:fca_across_dirs}, including both contranominal and ordinal motifs, suggests promising opportunities for analysis.
In accordance with expectations, the directories at the bottom of the graph,
which are considered the most specific in terms of topic categorization, predominantly contain subdirectories.
It is noteworthy that the only directory present at both the highest level of the directory hierarchy and the bottom node is the \texttt{Military} directory.
With respect to content, some directories exhibit intriguing thematic associations.
The \texttt{High-Tech Crime} directory, for instance, comprises topics related to explosives,
while the \texttt{Success Advice} directory additionally features content concerning chemicals.
The top node of the ordinal motif, which aggregates the incoming edges, introduces philosophical content.
Consequently, the content of directories positioned at the lowermost levels of the graph,
which predominantly comprise subdirectories,
encompasses a greater proportion of philosophical subjects in comparison to directories positioned
at more elevated levels within the graph and, in some cases, the directory hierarchy of the dataset.

Our approach \ac{approachname}
enables us to explore large datasets without the need for manual inspection of each document.
The hierarchical structure of the concept lattice offers insights
into the relationships between topics,
facilitating a deeper understanding of how different themes interconnect across directories.
However, it is important to note that while this approach offers valuable insights,
the visual representation may not be intuitive to all users,
particularly those unfamiliar with concept lattices or \ac{fca}.
The topics are represented by their IDs,
because the space is too limited to display 50 words per topic.

\begin{table}[H]
    \centering
    \caption{Subset of translated topic IDs and their top 5 descriptive words for the directory-topic concept lattice.}
    \label{tab:dir_topic_topics2terms}
    \resizebox{\textwidth}{!}{%
        \begin{tabular}{|
                >{\columncolor[HTML]{D4D4D4}}l |l|l|l|l|l|}
            \hline
            \rowcolor[HTML]{D4D4D4}
            \textbf{Topic ID} & \multicolumn{5}{c|}{\textbf{Topic Words}}                                                                   \\ \hline
            \textbf{34}       & gun                                       & weapons      & rifles          & pistol         & firearm       \\ \hline
            \textbf{36}       & terminological                            & nonverbal    & linguist        & codeword       & longtext      \\ \hline
            \textbf{54}       & microelectron                             & microkernels & microparticle   & microphysics   & microcosmic   \\ \hline
            \textbf{56}       & pythonpath                                & pythonmac    & python\_version & python\_object & pythonpowered \\ \hline
            \textbf{58}       & nuclear                                   & hazardous    & explosive       & nanotoxicity   & chernobyl     \\ \hline
            \textbf{94}       & project                                   & fundraiser   & donation        & volunteering   & charity       \\ \hline
        \end{tabular}%
    }
\end{table}

\subsection{Baselines}

\paragraph{Word clouds}

\begin{figure}[b]
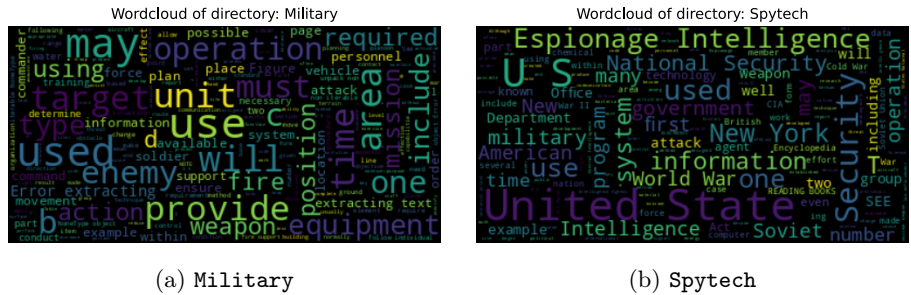

    \centering
    \begin{subfigure}{0.49\textwidth}
        \centering
        \includesvg[width=\textwidth]{images/wordclouds/wordcloud_Military}
        \caption{\texttt{Military}}
        \label{fig:wordcloud_military}
    \end{subfigure}
    \hfill
    \begin{subfigure}{0.49\textwidth}
        \centering
        \includesvg[width=\textwidth]{images/wordclouds/wordcloud_Spytech}
        \caption{\texttt{Spytech}}
        \label{fig:wordcloud_spytech}
    \end{subfigure}

    \caption{Wordcloud visualisations for selected directories. }
    \label{fig:wordclouds_overview}
\end{figure}

These word clouds not only provide an overview of the directory's content
based on the most frequent terms in the directory
but also allow identifying directories noteworthy of further inspection.
Intuitively, if the words of a word cloud do not align with the expected content of the directory name,
or if a word cloud includes terms of particular relevance,
one can consider the directory in further data mining.
While word clouds are intuitive and simple to interpret, they have limitations.
They only display the most frequent words, which may not fully represent the semantic relationships between terms.
Moreover, the quality of the word clouds is dependent on the
quality of the text extraction process (cf. Figure \ref{fig:extract_text}).
Furthermore, by inspection of the word clouds for the directories
\texttt{Military} and \texttt{Spytech},
respectively displayed in Figure \ref{fig:wordcloud_military} and \ref{fig:wordcloud_spytech},
it becomes evident that it is complicated to compare the content of different directories.

\paragraph{\acf{edr}}

\begin{figure}[h]
    \centering
    \includegraphics[width=.9\textwidth]{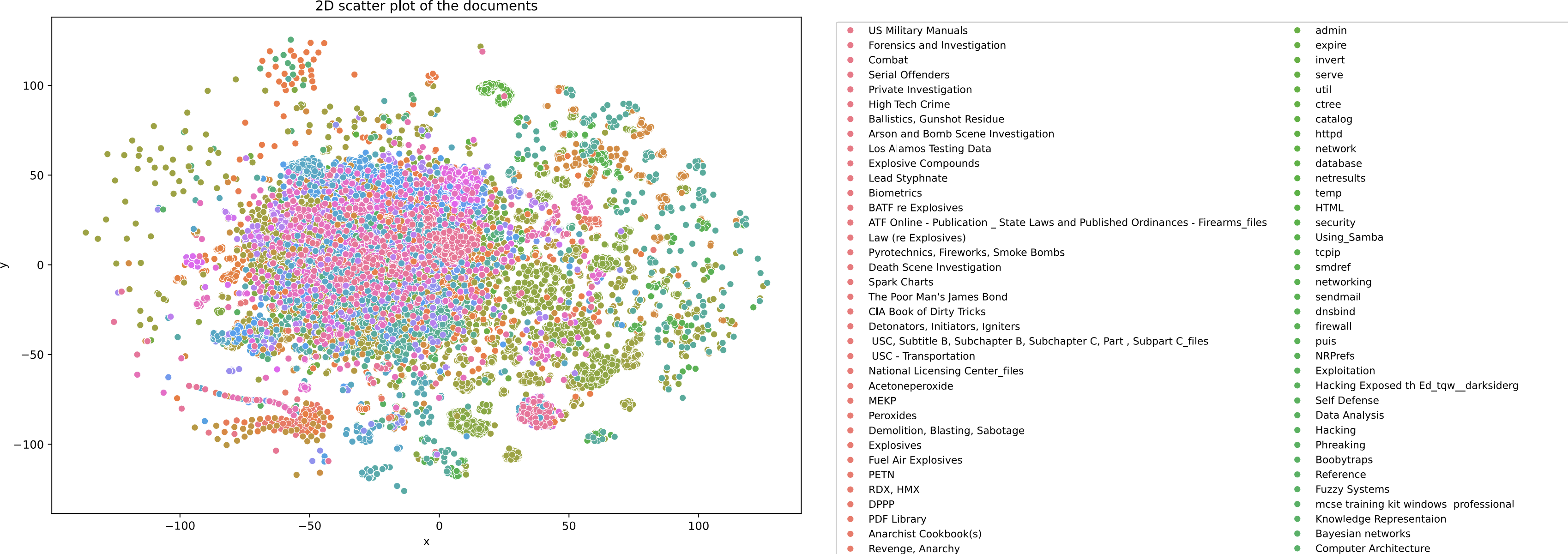}

    \caption{Scatter plot of document embeddings reduced to two dimensions using \ac{tsne}, colored by directory labels.}
    \label{fig:low_dim_dir_colour_tsne}
\end{figure}

Ideally, the dimensionality reduction algorithm should separate the clusters of documents
associated with different directories.
Although \ac{tsne} provides the best separation among the reducer algorithms,
the large number of directories results in too many classes to distinguish,
as shown in Figure \ref{fig:low_dim_dir_colour_tsne}.
Consequently, this approach is not well-suited for datasets of this size,
since it is difficult to extract meaningful clusters from the reduced embeddings,
as the colors become too similar.

\paragraph{\acf{cne}}

\begin{figure}[H]
    \centering
    \includegraphics[width=.9\textwidth]{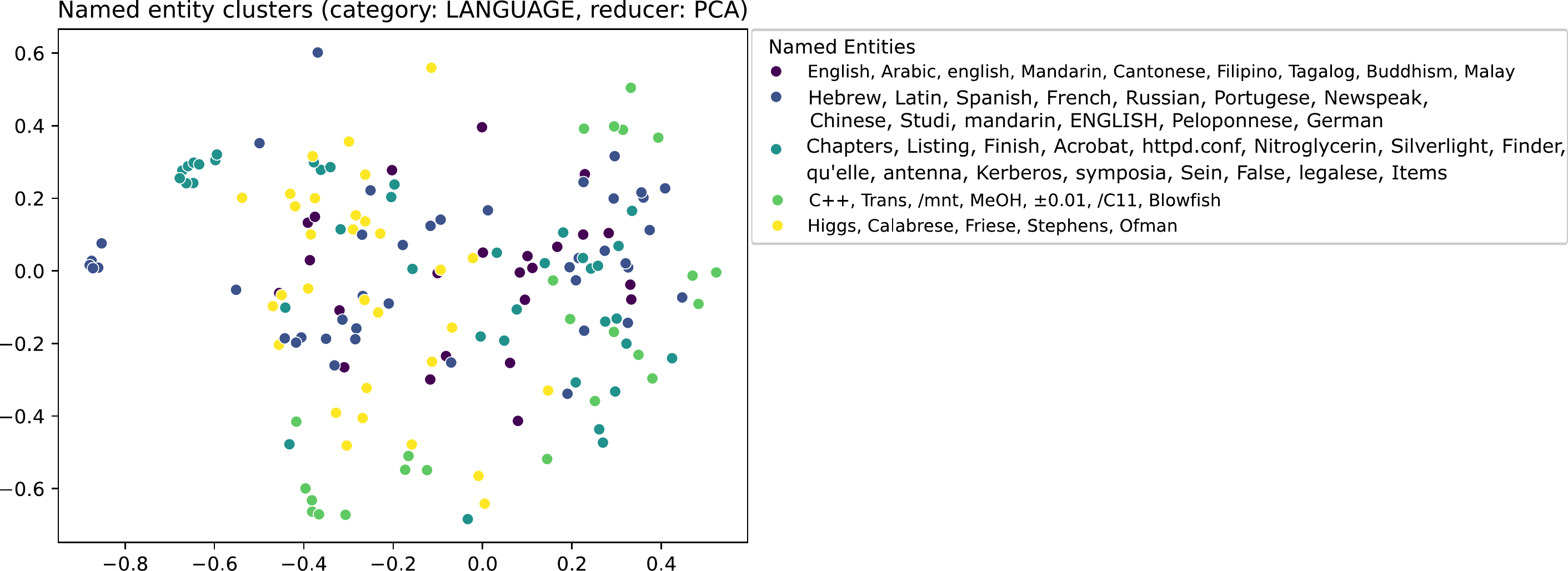}
    \caption{Clustering of \acp{ne} of the category \texttt{LANGUAGE}.}
    \label{fig:ne_clusterting_language}
\end{figure}

This baseline is based on entity categories,
in contrast to our \ac{approachname},
which relies on the hierarchical directory structure of the dataset.
A visualization of the clustering of the NE category \texttt{LANGUAGE}
is shown in Figure \ref{fig:ne_clusterting_language}.
It reveals that the approach effectively separates programming languages,
such as \textit{C++}, from spoken languages.
Most language
families\footnote{\url{https://www.omniglot.com/writing/langfam.htm?} (01.02.2025)}
are grouped together.
However, as with the \ac{ner} process, entities are often incorrectly categorized, e.g., \textit{False}.
Although some patterns in the class members can be observed,
neither the topological clusters nor
the clustering results provide clear or accurate information.
Overall, this approach does not yield meaningful insights into the data's structure.

\subsection{Empirical Comparison}
\label{sec:evaluation}

In this section, we compare different approaches for analyzing directory content,
assessing their strengths and limitations.
Our evaluation demonstrates that the proposed \ac{fca}-based method \ac{approachname}
offers superior insights into the dataset’s structure.

Word clouds provide an intuitive overview by highlighting frequently occurring terms,
enabling the identification of noteworthy directories.
However, they fail to capture semantic relationships and
depend heavily on the text extraction quality.
Additionally, comparing multiple directories using word clouds is challenging
due to their unstructured nature.
In contrast, embedding-based visualization techniques allows directly comparing directories
based on their content.
Not only does the quality of the reducers affect the results,
but the large number of directories also complicates the interpretation of the visualization.
This approach is therefore ineffective for large-scale datasets.
\ac{ner}-based clustering of category entities suffers from frequent misclassification.
Moreover, the resulting clusters neither exhibit clear patterns nor provide meaningful structural insights into the content of directories,
making this method unreliable for data analysis.
Our proposed \ac{fca}-based approach enables a hierarchical exploration of directory topics
without requiring manual document inspection.
Although the visualization is less intuitive for non-experts,
this method remains the most effective for uncovering the relationships between directory topics.
Among the evaluated methods, \ac{approachname} proves to be the most effective for structured data exploration.

\section{Conclusion and Future Work}
\label{sec:future_work}

In this paper, we proposed a method for analyzing structures in an unknown dataset of diverse content, and compared it to three baseline methods.
In the case study, we found that our approach \ac{approachname} is better suited for exploring and visualizing topics across multiple directories
than the baseline techniques.
While word clouds offer a quick overview, and embedding-based methods attempt clustering,
neither approach provides sufficient clarity for the large dataset of the case study.
\ac{ner} clustering is unreliable due to misclassification.
In contrast, \ac{fca} enables hierarchical and systematic topic analysis,
making it the preferred method despite requiring expertise for interpretation.

However, our approach has some limitations. Visualizing topic names directly in the concept lattice leads to visual clutter, while using only topic IDs makes the lattice hard to interpret for a human.
To solve this, one may experiment with interactive versions where mouseover effects reveal topic details.
Additionally, the quality of image captioning and topic modeling might limit the expressiveness of the computed concept lattices.

Currently, \ac{approachname}'s results do not provide any insight into the dataset's directory hierarchy.
However, this structural information could be highly valuable when considered alongside
the topic hierarchy of the directories.
Therefore, future work could explore integrating both sources of information to enhance the analysis.

In this work, we employed Top2Vec \cite{top2vec_2020} as the topic modeling technique.
While newer models, such as C-Top2Vec \cite{top2vec_2024}, have been introduced,
we opted for Top2Vec due to its ability to represent documents using a single-vector format.
This approach allows for topic aggregation by counting the occurrences of topics across the documents in a directory.
In contrast, C-Top2Vec supports multi-vector document representation,
which would provide a more granular description of individual documents and the dataset as a whole.
Future work could explore and evaluate the performance of C-Top2Vec in comparison to Top2Vec for this purpose.

\begin{acronym}

    \acro{fca}[FCA]{Formal Concept Analysis}
    \acro{dataset}[ETYNTKE]{Everything You Need To Know Ever}
    \acro{nlp}[NLP]{Natural Language Processing}
    \acro{git}[GIT]{Generative Image-to-text Transformer}
    \acro{pdf}[PDF]{Portable Document Format}
    \acro{ner}[NER]{Named Entity Recognition}
    \acro{ne}[NE]{Named Entity}
    \acro{sbert}[SBERT]{Sentence-BERT}
    \acro{use}[USE]{Universal Sentence Encoder}
    \acro{es}[ES]{Elasticsearch}
    \acro{umap}[UMAP]{Uniform Manifold Approximation and Projection for Dimension Reduction}
    \acro{tsne}[t-SNE]{t-Distributed Stochastic Neighbor Embedding}
    \acro{pca}[PCA]{Principal Component Analysis}
    \acro{hdbscan}[HDBSCAN]{Hierarchical Density-Based Spatial Clustering of Applications with Noise}
    \acro{bow}[BoW]{Bag of Words}
    \acro{tfidf}[TF-IDF]{Term Frequency-Inverse Document Frequency}
    \acro{approachname}[FAT-CAT]{FCA-based Aggregation for Topics using Conceptual Analysis and Taxonomies}
    \acro{cne}[CNE]{Clustering Named Entities}
    \acro{edr}[EDR]{Embedding - Dimension Reduction}
    \acro{hlda}[hLDA]{Hierarchical Latent Dirichlet Allocation}

\end{acronym}

\bibliographystyle{splncs04}
\bibliography{bibliography}

\newpage

\end{document}